\documentclass{article}

\usepackage[preprint]{corl_2026} 
\usepackage{hyperref}

\usepackage{orcidlink}
\usepackage{booktabs}
\usepackage{multirow}
\usepackage{graphicx}
\usepackage{xcolor}
\usepackage{xspace}
\usepackage{float}
\usepackage{enumitem}

\newcommand{\name}{\texttt{AnySlot}\xspace}

\newcommand{\stitle}[1]{\noindent{\bf #1\/}}
\usepackage{graphicx}
\usepackage{booktabs}
\usepackage{pifont}
\usepackage{multirow}
\usepackage{array}
\usepackage{bbm}
\usepackage{xcolor}
\usepackage[table]{xcolor}
\usepackage{array}

\usepackage[accsupp]{axessibility}  

\title{AnySlot: Goal-Conditioned Vision-Language-Action Policy for Zero-Shot Slot-Level Placement}

\author{
  Zhaofeng Hu\\
  Stony Brook University\\
  Stony Brook, USA\\
  \texttt{zhaofeng.hu@stonybrook.edu} \\
  \And
  Sifan Zhou \\
  Carnegie Mellon University \\
  Pittsburgh, USA \\
\texttt{sifanjay@gmail.com,} \\
  \And
  Qinbo Zhang \\
  Stony Brook University \\
  Stony Brook, USA \\
  \texttt{qinbo.zhang@stonybrook.edu} \\
  \And
  Rongtao Xu \\
  MBZUAI \\
  Abu Dhabi, UAE \\
  \texttt{xurongtao2022@gmail.com} \\
  \And
  Qi Su \\
  Peking University \\
  Beijing, China \\
\texttt{qiisuu@stu.pku.edu.cn} \\
  \And
  Jorge Mendez-Mendez \\
  Stony Brook University \\
  Stony Brook, USA \\
  \texttt{jorge.mendezmendez@stonybrook.edu} \\
  \And
  Ci-Jyun Liang \\
  Stony Brook University \\
  Stony Brook, USA \\
  \texttt{ci-jyun.liang@stonybrook.edu} \\
}

\begin{document}
\maketitle


\begin{figure}[H]
\centering
\includegraphics[width=0.75\linewidth]{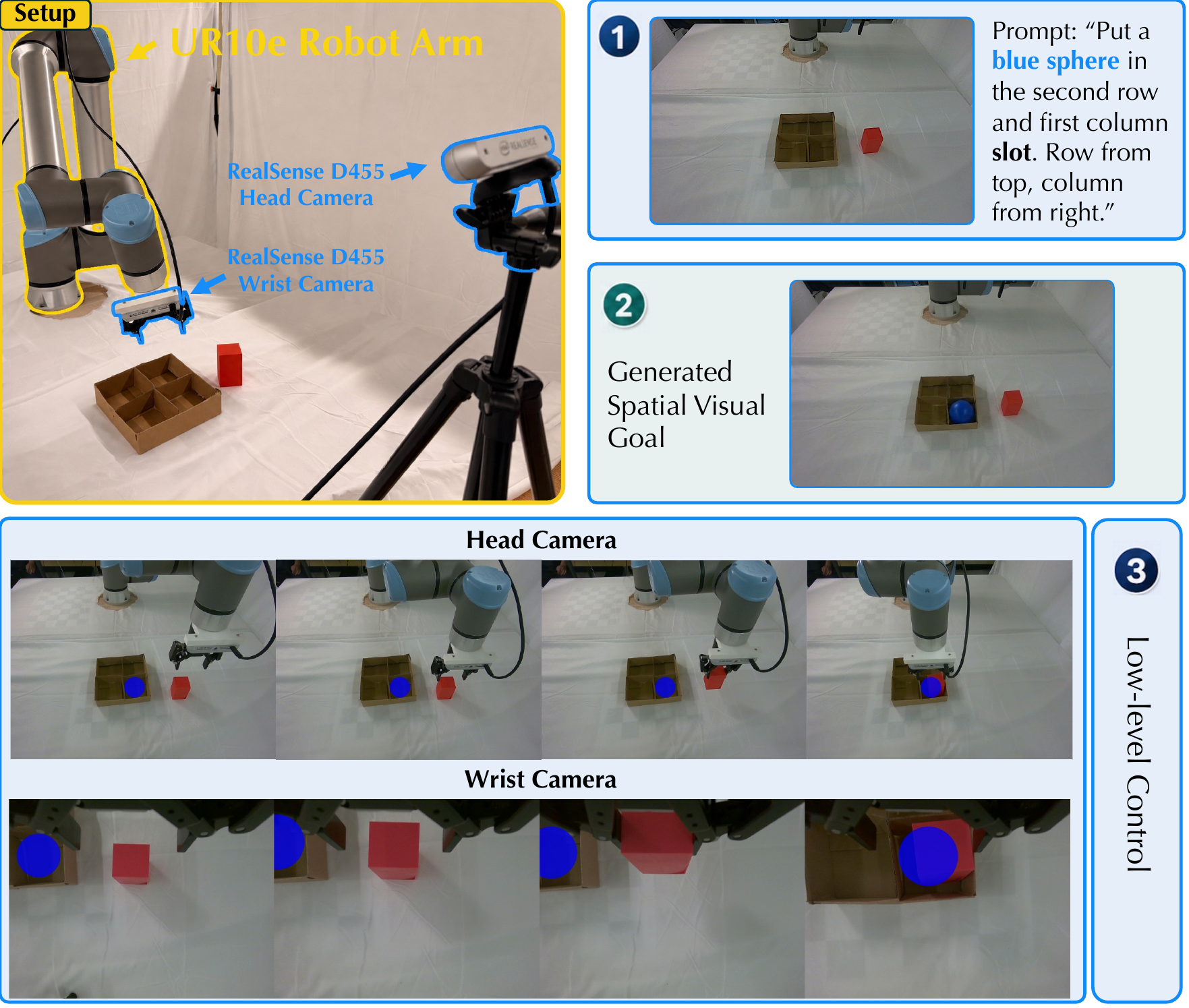}
\caption{
\textbf{Real-world goal-guided slot-level placement.}
(1) Given a language instruction, the robot captures a head-camera RGB-D observation.
(2) The image-generation module renders a spatial visual goal at the instructed slot.
(3) The goal is reconstructed and projected into head and wrist camera views to guide low-level manipulation, enabling successful placement on a real UR10e platform.
\name{} demonstrates real-world generalization and robustness.
}

\label{fig:real_goal}
\end{figure}

\begin{abstract}
Vision-Language-Action (VLA) policies have emerged as a versatile paradigm for generalist robotic manipulation. However, precise object placement under compositional language remains challenging for end-to-end VLA policies. Slot-level placement requires reliable slot grounding and centimeter-level geometric precision. To this end, we propose \name, a framework that reduces compositional complexity by introducing an explicit spatial visual goal between language grounding and control. \name converts language into a visual goal by rendering a spatial marker at the intended slot, then executes this goal with a goal-conditioned VLA policy. This hierarchical design decouples high-level slot selection from low-level execution, improving semantic accuracy and spatial robustness. Furthermore, recognizing the lack of benchmarks for such precision-demanding tasks, we introduce \textbf{SlotBench}, a structured simulation benchmark with nine task categories for evaluating spatial reasoning in slot-level placement. Extensive experiments show that \name significantly outperforms flat VLA baselines and modular grounding methods in zero-shot slot-level placement.
\end{abstract}

\keywords{Vision-Language-Action, Robot Manipulation, Slot-level Placement.} 


\section{Introduction}

Recent robotic policies, including flat Vision-Language-Action (VLA)~\cite{zitkovich2023rt,kim24openvla,black2024pi_0,intelligence2025pi_}) and modular Vision-Language Model (VLM)~\cite{karamcheti2024prismatic,beyer2024paligemma,radford2021learning,steiner2024paligemma,deitke2025molmo}) based policies~\cite{KITE2023,huang2025roboground,shi2025hi,yuan2025seeing,huang2025thinkact,li2025hamster}, have made impressive progress on language-conditioned manipulation. However, many practical placement tasks require more than moving an object to an approximately correct region. In settings such as assembly, sorting, and tool organization, a robot must place an object into an exact slot among many visually similar candidates, which we refer to as \emph{slot-level placement}~\cite{shan2025slot}. This task requires both semantic correctness, i.e., grounding language to the intended slot, and geometric correctness, i.e., executing precise placement within tight slot boundaries. It becomes especially challenging when the target is specified by relational, ordinal, functional, or commonsense instructions, such as ``place it into the most stable compartment'' or ``avoid the lower-left cells.'' Despite recent progress, existing language-conditioned robotic policies remain insufficient for such fine-grained placement.

\begin{figure}[!t]
\centering
\includegraphics[,
width=0.90\linewidth,
    trim=0cm 0.5cm 0cm 0cm,
    clip
]{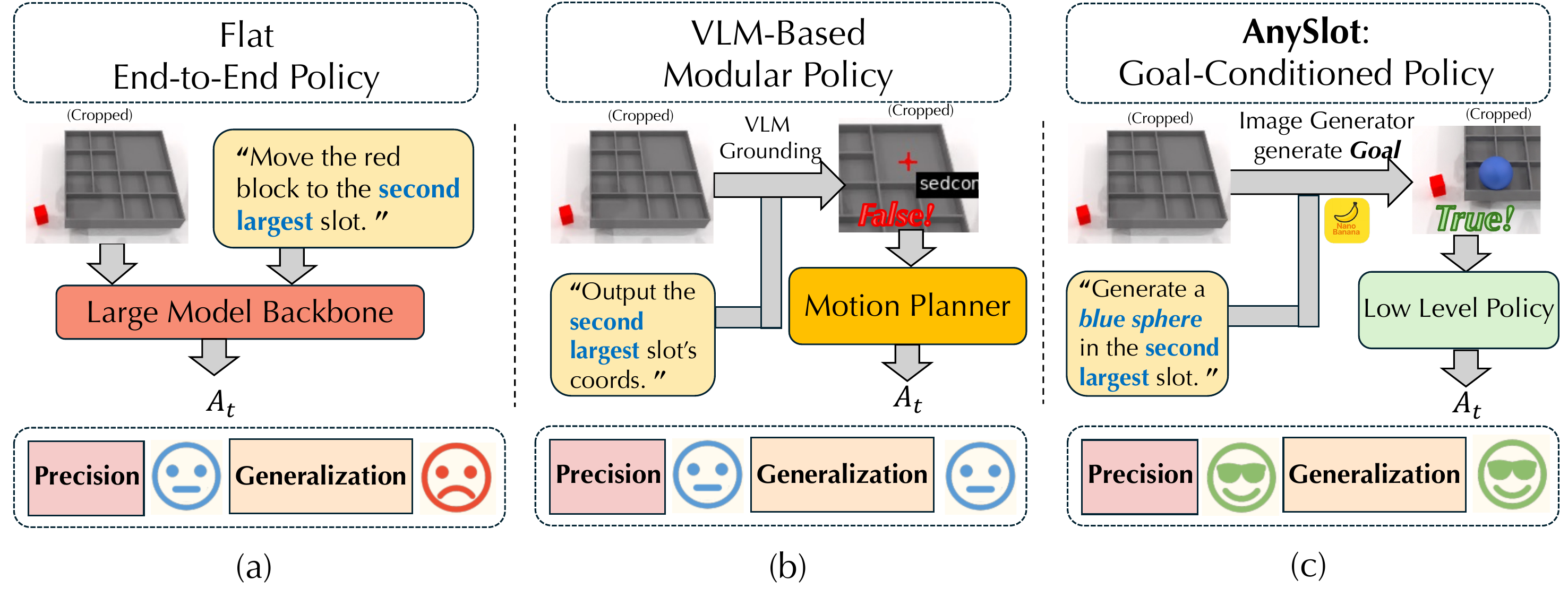}
\caption{\textbf{Overview of flat (a), modular (b), and (c) our goal-conditioned policy.} For slot-level placement tasks, our method achieves superior zero-shot generalization by transforming language into explicit spatial visual goals for reliable execution.}

\label{teaser}
\end{figure}

As shown in Fig.~\ref{teaser} (a), \textbf{Flat end-to-end policies} directly map multi-modal observations and language instructions to control actions. However, their VLM backbones remain limited in compositional spatial reasoning~\cite{shukor2025scaling}, and the policy must implicitly perform slot grounding while generating low-level actions. This entangles slot selection and motor control in a single output space, making it difficult to maintain both semantic correctness and geometric precision under dense, unseen slot layouts. To reduce policy complexity, \textbf{VLM-based modular policies} adopt a two-stage design that first grounds the instruction with a VLM and then executes placement through a downstream controller, as shown in Fig.~\ref{teaser}(b). However, although VLMs exhibit certain spatial reasoning ability~\cite{yang2025thinking}, their grounding outputs are often coarse, token-based, or represented as a single coordinate. Such representations make precise localization unreliable and discard local geometric context, such as slot boundaries, which is essential for stable alignment under unseen layouts. We provide a broader discussion of related VLA policies, modular robotic systems, and placement methods in Appendix.

To address these challenges, we propose \name, a two-stage framework that converts language into a spatial visual goal by an image generation model and uses a VLA policy for precise placement. Our key insight is that the interface between high-level reasoning and low-level control should be a spatially coherent visual goal in the robot's observation space. Instead of relying on a VLM to regress a single coordinate, \name uses an image generation model as a high-level grounding module to render a visual goal directly at the intended slot. This turns language grounding into a visually explicit and spatially coherent goal representation that can be consumed by a low-level policy with strong visuomotor priors. Conditioned on the generated goal, a low-level  VLA policy executes pick-and-place, leveraging learned robustness to perception noise and contact-rich manipulation.

To systematically evaluate slot-level grounding and manipulation, we introduce SlotBench, a benchmark for structured slot-level reasoning in robotic placement comprising nine challenge tasks, designed to probe zero-shot generalization beyond seen layouts and instructions. SlotBench covers ordinal reasoning, size and height comparison, distance reasoning, compositional relations, logical negation, vague language, affordance reasoning, and world knowledge. Models must succeed on novel slot configurations and previously unseen instructions without task-specific finetuning. As shown in Tab.~\ref{tab:main_exp}, existing flat and modular pipelines struggle to solve these tasks reliably, often failing on most categories, while \name achieves nearly \textbf{90\%} average success, demonstrating the effectiveness of explicit visual goal construction for precise slot-level placement.

To summarize, we make three contributions:
\begin{enumerate}[leftmargin=*, itemsep=0pt, topsep=2pt]
    \item We identify slot-level placement as a practical importance yet underexplored VLA problem requiring exact language grounding and precise execution under tight geometric constraints.
    \item We propose \textbf{\name}, a two-stage framework that converts language instructions into explicit spatial visual goals, providing a structured reasoning-control interface for slot-level placement.
  \item We introduce SlotBench, a diverse benchmark for evaluating slot-level spatial reasoning and language grounding under zero-shot generalization to unseen layouts and instructions.
\end{enumerate}

\section{Preliminaries}
\label{sec:prelim}

\stitle{Policy formulation.}
At time step \(t\), a robot policy observes \(o_t=[I_t^{1:m},q_t,\ell_t]\), where \(I_t^{1:m}\) are multi-view images, \(q_t\) is proprioception, and \(\ell_t\) is the language instruction. 
It outputs either an action \(a_t\) or an action chunk \(A_t=[a_t,\dots,a_{t+H-1}]\) for the next \(H\) steps~\cite{zhao2023learning}. Then action generation is modeled as the conditional distribution \(p(A_t\mid o_t)\), denoted by \(\pi_\theta(\cdot\mid o_t)\) for a learned parameterization.

\stitle{Vision-Language-Action policies.}
Vision-Language-Action (VLA) policies extend Vision-Language Models (VLMs) from text generation to robot action generation.
Given \((I,\ell)\), a VLM models \(p(\ell' \mid I,\ell)\), typically autoregressively, while a VLA conditions on robot observations \(o_t\) and models \(p(A_t \mid o_t)\) or \(\pi_\theta(\cdot \mid o_t)\).
Existing VLAs commonly use either \textbf{\ding{182} tokenized decoding}, where actions are discretized and predicted sequentially, e.g.,
\(p(A_t \mid o_t) \approx p(\mathrm{tok}(A_t)\mid I_t^{1:m}, q_t, \ell_t)\)~\cite{kim24openvla,zitkovich2023rt}, or \textbf{\ding{183} continuous generative modeling}, where actions are generated as continuous variables, e.g.,
\(A_t = g_\theta(\varepsilon;\, o_t)\), \(\varepsilon \sim \mathcal{N}(0,I)\)~\cite{black2024pi_0,intelligence2025pi_}.
Since slot-level placement requires precise continuous control, we build on a continuous generative VLA, \(\pi_{0.5}\), for multi-modal execution.

\stitle{Problem formulation: slot-level placement.} 
We study slot-level object placement~\cite{shan2025slot} under compositional language instructions in zero-shot environments. 
Given candidate slots \(\mathcal{S}=\{s_i\}_{i=1}^{N}\), the instruction \(\ell_t\) specifies a target slot \(s^\star\in\mathcal{S}\). 
The task requires \textbf{\ding{182} slot selection:} identifying \(s^\star\), and \textbf{\ding{183} precision placement:} placing the object with sufficient geometric accuracy. 
Here, compositional language denotes instructions that combine spatial, relational, comparative, negation, affordance, or commonsense constraints, while zero-shot environments contain unseen language compositions and slot layouts without task-specific fine-tuning. 
Directly mapping such instructions to low-level actions is brittle, since small grounding errors can cause placement failures. 
To reduce compositional complexity, we introduce an intermediate spatial visual goal \(G_t\), derived from language and perception via an image generation model, and model execution as \(p(A_t \mid o_t, G_t)\).
	

\section{Method}

\label{sec:method}
\begin{figure}[!t]
\centering
\includegraphics[width=0.86\linewidth]{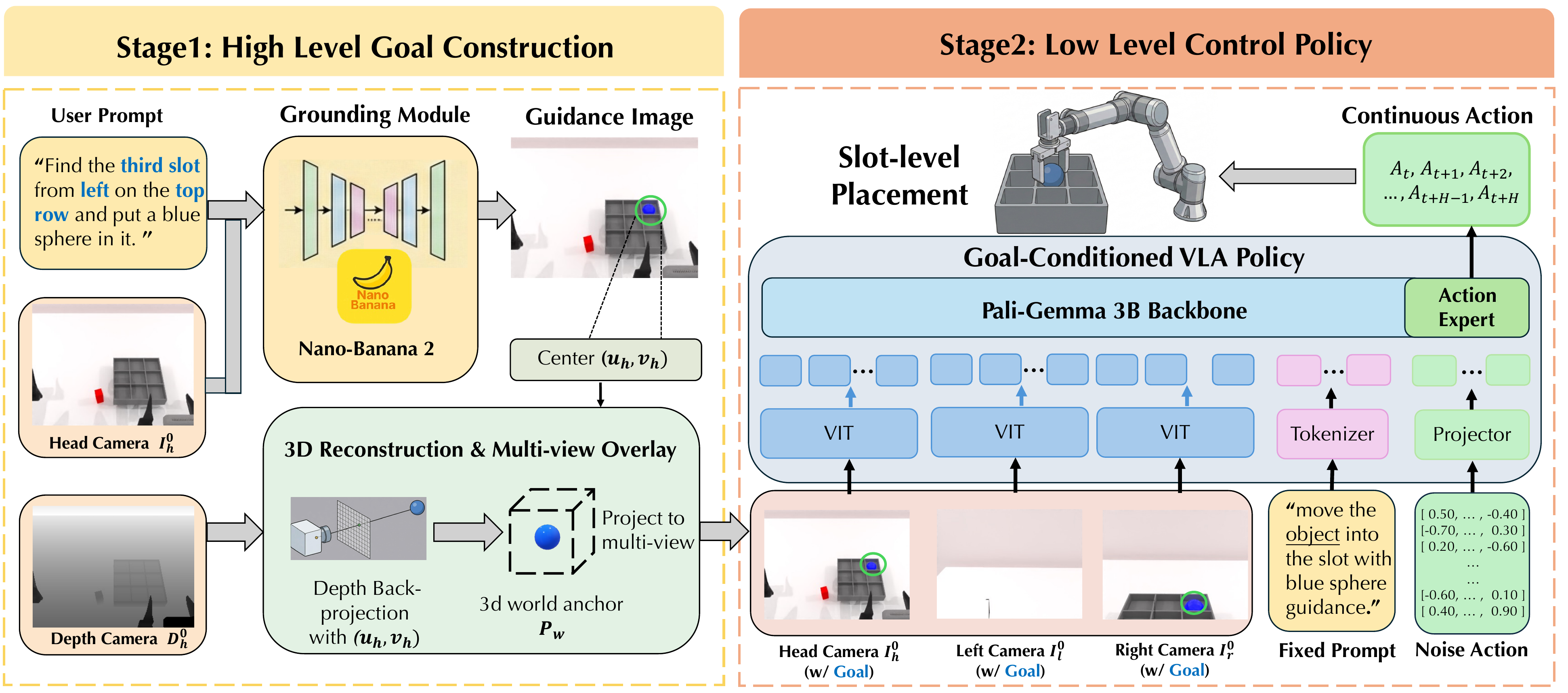}
\caption{\textbf{\name overview.}
We formulate slot-level placement as goal-conditioned control. \textbf{High-level goal construction} uses the Nano-Banana image generator to render a blue-sphere goal from the language prompt, lifting it to a view-consistent multi-view overlay via depth and camera calibration. \textbf{Low-level control} uses a goal-conditioned VLA policy (\(\pi_{0.5}\)) with a PaliGemma-3B backbone and action expert, mapping goal-augmented multi-view observations to continuous actions \(A_t\).}
\label{fig:framework}

\end{figure}

\subsection{Method Overview}
\label{sec:method_overview_text}

Fig.~\ref{fig:framework} provides an overview of \name. At time step \(t\), the robot receives the observation \(o_t = [I_t^{1:m}, q_t, \ell_t]\). 
The key challenge is to satisfy compositional language constraints while maintaining geometric precision under unseen layouts. To reduce this compositional complexity, an explicit intermediate goal variable \(G_t\) is introduced and the policy is factored through it:
\begin{equation}
p(A_t \mid o_t) \;=\; \int p(A_t \mid o_t, G_t)\, p(G_t \mid o_t)\, dG_t,
\label{eq:anyslot_factorization}
\end{equation}
where \(G_t\) denotes the intermediate spatial visual goal derived from language and perception, \(p(G_t \mid o_t)\) performs goal inference (grounding) and \(p(A_t \mid o_t, G_t)\) performs goal-conditioned execution. 

In \name, the goal inference term \(p(G_t \mid o_t)\) is instantiated by a generation-based grounding module that converts language into an explicit visual goal. Concretely, an out-of-the-box (e.g., Nano-Banana) image generation model renders a colored sphere marker in the head image, and the marker is lifted to a world-space anchor using the aligned depth map and camera calibration. This anchor is then projected back into all camera streams to form view-consistent overlays, which constitute the visual goal \(G_t\) consumed by the policy (Sec.~\ref{sec:goal_construction}). 

The execution term \(p(A_t \mid o_t, {G}_t)\) is instantiated with a continuous generative VLA policy (e.g., \(\pi_{0.5}\)), which conditions on multi-view observations augmented with the rendered sphere overlays (Sec.~\ref{sec:low_level_policy}). To isolate high-level language grounding from low-level control, the low-level instruction is kept fixed, and spatial intent, which slot to use, is provided exclusively through the visual goal \(G_t\). This design bridges structured language grounding and precise execution while remaining compatible with current multi-camera VLA policies.

\subsection{Stage 1: High-level Goal Construction Module}
\label{sec:goal_construction}

The goal construction module produces an explicit spatial visual goal \(G_t\) from the head observation and language instruction, and renders it consistently across views for downstream multi-view policies (e.g., \(\pi_{0.5}\)). 
We consider a multi-view setup with a fixed head RGB-D camera \(c_h\) and a wrist camera \(c_w(t)\) mounted on the end-effector. 
Let \(K_h,K_w\) denote camera intrinsics, and let \(^{W}\!T_{c_h}\) denote the calibrated head-camera pose in a shared world frame \(W\). 
At the beginning of an episode, the head RGB image \(I_h^0\) and depth map \(D_h^0\) are captured. 
An out-of-the-box image generation model \(g_\phi\) produces an edited guidance image \(\tilde{I}_h^0=g_\phi(I_h^0,\ell)\), where a blue sphere marker indicates the instructed slot. 
We extract the marker center \((u_h,v_h)=\mathcal{C}(\tilde{I}_h^0)\) using HSV segmentation, morphological filtering, and ellipse fitting.

\begin{figure}[t]
\centering
\includegraphics[,
width=1\linewidth,
    trim=0cm 0.2cm 0cm 0.2cm,
    clip
]{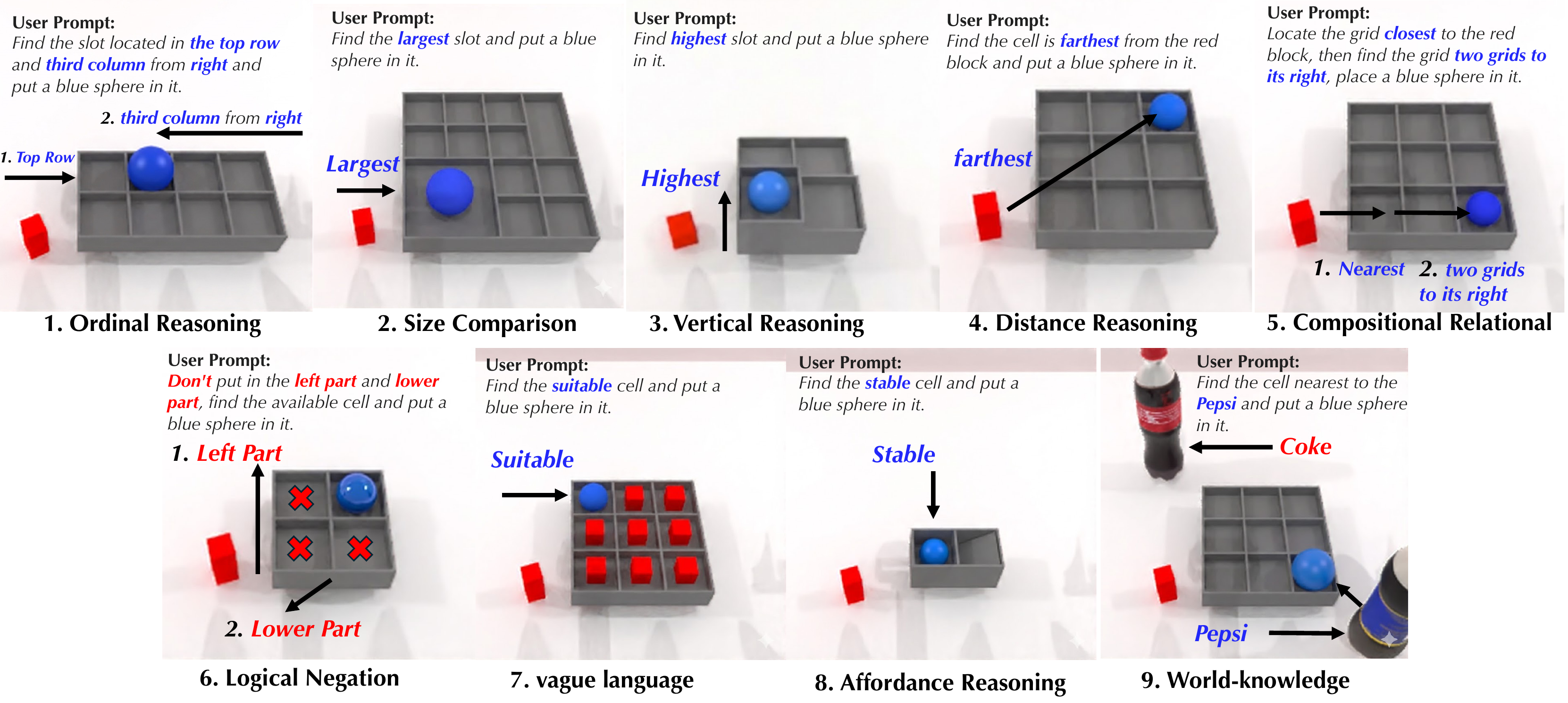}
\caption{\textbf{SlotBench.} A slot-level benchmark with nine challenges for spatial reasoning and language grounding in manipulation.
}
\label{fig:slotbench}

\end{figure}

Given the marker depth \(z_h=D_h^0(u_h,v_h)\), the 2D marker is back-projected into the head-camera frame and transformed into a world-space anchor:
\begin{equation}
\mathbf{p}^{c_h}=z_hK_h^{-1}[u_h,v_h,1]^\top,\qquad
\mathbf{p}^{W}=\,^{W}\!T_{c_h}[(\mathbf{p}^{c_h})^\top,1]^\top .
\end{equation}
This single anchor is then re-projected into each camera stream to generate view-consistent goal overlays. 
The wrist-camera pose is obtained from forward kinematics and hand--eye calibration as \(^{W}\!T_{c_w}(t)=\,^{W}\!T_{ee}(q_t)\,^{ee}\!T_{c_w}\). 
For any camera \(c\in\{c_h,c_w(t)\}\), the world anchor is projected as
\begin{equation}
\tilde{\mathbf{p}}^{c}=(^{W}\!T_{c})^{-1}[(\mathbf{p}^{W})^\top,1]^\top,\qquad
\lambda[u,v,1]^\top=K_c[X^c,Y^c,Z^c]^\top,
\end{equation}
where \(\tilde{\mathbf{p}}^{c}=[X^c,Y^c,Z^c,1]^\top\). 
A blue sphere marker is rendered at \((u,v)\) in each view, with pixel radius \(r_{\mathrm{px}}\approx f_x r/Z^c\), producing overlaid observations \(\{I_h^{\mathrm{ov}}(t),I_w^{\mathrm{ov}}(t)\}\) consumed by the policy. 
By deriving all overlays from the same world-space anchor, \(G_t\) provides an aligned multi-view goal cue compatible with current multi-camera VLAs.




\subsection{Stage 2: Low-Level Goal-Conditioned Policy}
\label{sec:low_level_policy}

Given the observation $o_t$ and the spatial visual goal $G_t$ constructed in Sec.~\ref{sec:goal_construction}, execution is modeled as a goal-conditioned action distribution $p(A_t \mid o_t, G_t)$. 
In \name, this distribution is instantiated with $\pi_{0.5}$, a continuous generative VLA that uses a PaliGemma-3B vision-language backbone and a flow-matching action expert. 
Concretely, $\pi_{0.5}$ generates an action chunk as $A_t=g_\theta(\varepsilon;\,o_t,G_t)$ with $\varepsilon\sim\mathcal{N}(0,I)$, where $G_t$ is represented as a rendered sphere overlay in the policy input. 

We fine-tune $\pi_{0.5}$ on a synthetic goal-guided dataset $\mathcal{D}_{\mathrm{syn}}$ collected in RoboTwin~\cite{Xiang_2020_SAPIEN,chen2025robotwin}, which have 450 episodes contain multi-view observations, robot states $(I_t^{1:m},q_t)$, continuous action targets, and an oracle visual goal $G_t^\star$ rendered at the intended slot. 
This teaches the policy to ground and track the blue sphere for precise slot-level execution. 
Implementation details are provided in Sec.~\ref{sec:experiments}.

To decouple high-level language grounding from low-level control, the low-level instruction is fixed and does not require language-based spatial reasoning. 
The policy is prompted with the instruction template ``move the \(\underline{object}\) into the slot with blue sphere guidance'', where \(\underline{object}\) specifies only the object category. 
Slot selection is specified solely through $G_t$, and the low-level policy focuses on goal-conditioned execution rather than instruction-level slot reasoning.

\section{SlotBench: A Slot-Level Placement Benchmark}
\label{sec:slotbench}

\stitle{Overview.} 
To evaluate slot-level reasoning for robotic placement, we introduce \textbf{SlotBench}, a simulation benchmark built in RoboTwin~\cite{Xiang_2020_SAPIEN,chen2025robotwin}. SlotBench comprises nine task categories that test distinct forms of structured spatial reasoning for slot-level placement. Fig.~\ref{fig:slotbench} provides a visualization of SlotBench.
Unlike existing manipulation benchmarks that evaluate object-level goal completion, SlotBench isolates fine-grained slot selection under compositional language. To our knowledge, no benchmark specifically evaluates slot-level placement with explicit slot-level reasoning. Detailed task definitions and prompt examples are provided in the appendix.


\stitle{Benchmark composition.}
SlotBench contains nine task categories with diverse scene configurations. 
Each task includes at least five randomized scenes with different tray geometries, resulting in at least 45 unique scene layouts in total. 
For each scene, we generate multiple language instructions (at least five variants) that describe the target slot for both the high-level module and the flat policy. Compared with common VLA benchmarks\cite{liu2023libero,mees2022calvin,li2024simpler}, which emphasize object-level completion or continuous pose reaching, SlotBench targets discrete slot-level placement that requires structured spatial reasoning.
Small changes in language or scene geometry can switch the target slot, making SlotBench well-suited for evaluating reasoning-driven manipulation and compositional generalization.


\section{Experimental Evaluation}
\label{sec:experiments}

In this section, we study slot-level placement tasks that couple stringent geometric constraints with compositional language guidance in a zero-shot setting. We compare our \name against flat VLA baselines and prior modular approaches, which also decompose grounding and execution into separate stages. Our experiments are designed to \textbf{\ding{182}} evaluate slot-level placement under compositional language and unseen configurations, \textbf{\ding{183}} compare against flat VLA and modular baselines, and \textbf{\ding{184}} analyze the roles of goal construction and low-level control through ablations.

%
%


\begin{figure}[h]

\centering
\includegraphics[width=0.98\linewidth]{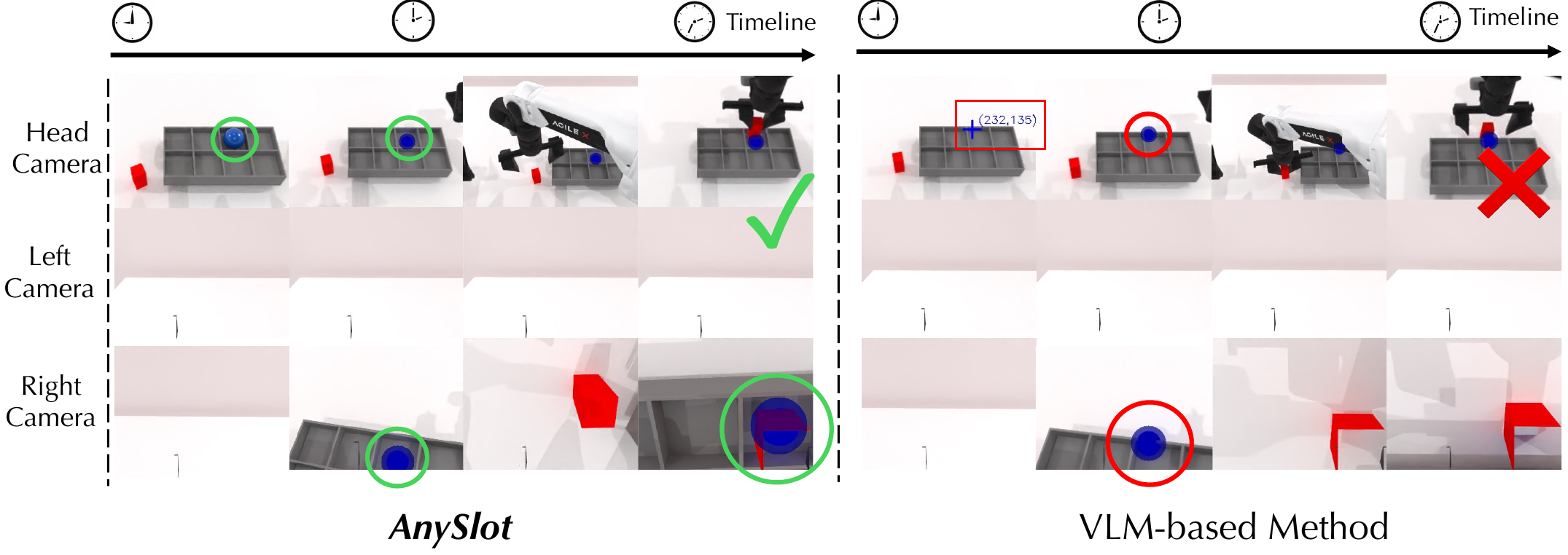}
\caption{Comparison between \name\ and a VLM-based method. 
\name\ grounds the target slot and succeeds, 
while the VLM-based method mislocalizes the target and fails.}

\label{fig:exp}
\end{figure}

\stitle{Evaluation benchmark.}
We evaluate on \textbf{SlotBench}, a zero-shot benchmark for slot-level placement under compositional language. It contains nine task categories (Sec.~\ref{sec:slotbench}) with randomized tray geometries, distractors, and object poses. Full task specifications are provided in the appendix.


\stitle{Baselines.} We compare with representative flat and modular policies. \textbf{\ding{182} Flat policy.} End-to-end policies directly map observations and language to actions, without explicit spatial visual goals. They are trained on a standard dataset \(\mathcal{D}_{\mathrm{std}}\) with 450 slot-level placement episodes using direct row--column language instructions. \textbf{\ding{183} Modular policy.} Two-stage methods first predict placement targets using grounding pipelines, such as VLM-based coordinate prediction, and then execute placement using motion planning or learned control. \textbf{\ding{184} Oracle high-level.} An expert-human oracle provides the correct target slot, measuring the low-level policy performance under perfect high-level grounding. Details of all baselines are provided in the appendix.

\stitle{Metrics.} We use three metrics. \textbf{\ding{182} Success Rate (SR)} measures successful placement into the ground-truth target slot. \textbf{\ding{183} Instruction Accuracy (IA)} measures whether the predicted target center is within \(0.02\,\mathrm{m}\) of the ground-truth slot center, reflecting SlotBench's tight geometry and centimeter-level precision requirement. \textbf{\ding{184} Coarse Accuracy (CA)} measures whether the predicted center falls inside the target slot region, and is reported in ablations to analyze grounding quality.


\stitle{Implementation Details.} Our default baseline uses out-of-box Nano-Banana 2 as the high-level module and \(\pi_{0.5}\)  as the low-level policy. We fully fine-tune \(\pi_{0.5}\)  with {\(\mathcal{D}_{\mathrm{syn}}\)} for 20{,}000 steps using batch size 64 on a single NVIDIA H200 GPU. More details are summarized in Appendix.

\setlength{\arrayrulewidth}{0.5pt} 
\setlength{\tabcolsep}{3pt}
\renewcommand\arraystretch{1.4}
\begin{table*}[!t]
    \tiny
    \caption{\textbf{Performance comparison of different methods on the SlotBench benchmark.} Each entry reports SR (Success Rate; higher is better, \%) and IA (Instruction Accuracy; higher is better, \%). The yellow rows denote our \name, while the blue rows indicate the oracle upper bound.}
    \centering
    \scalebox{0.9}{
    \begin{tabular}{c c |
        p{0.45cm}<{\centering} p{0.45cm}<{\centering}|
        p{0.45cm}<{\centering} p{0.45cm}<{\centering}|
        p{0.45cm}<{\centering} p{0.45cm}<{\centering}|
        p{0.45cm}<{\centering} p{0.45cm}<{\centering}|
        p{0.45cm}<{\centering} p{0.45cm}<{\centering}|
        p{0.45cm}<{\centering} p{0.45cm}<{\centering}|
        p{0.45cm}<{\centering} p{0.45cm}<{\centering}|
        p{0.45cm}<{\centering} p{0.45cm}<{\centering}|
        p{0.45cm}<{\centering} p{0.45cm}<{\centering}}
        \toprule[1.2pt]
        \multirow{2}{*}{\textbf{Group}} & \multirow{2}{*}{\textbf{Method}} 
        & \multicolumn{2}{c|}{\textbf{Ord.}} & \multicolumn{2}{c|}{\textbf{Size}} 
        & \multicolumn{2}{c|}{\textbf{Hgt.}} & \multicolumn{2}{c|}{\textbf{Dist.}} 
        & \multicolumn{2}{c|}{\textbf{Comp.}} & \multicolumn{2}{c|}{\textbf{Neg.}} 
        & \multicolumn{2}{c|}{\textbf{Vague}} & \multicolumn{2}{c|}{\textbf{Aff.}} 
        & \multicolumn{2}{c}{\textbf{World}} \\
        \cmidrule{3-20}
        & & SR & IA & SR & IA & SR & IA & SR & IA & SR & IA & SR & IA & SR & IA & SR & IA & SR & IA \\
        \cmidrule{1-20}
        \multirow{4}{*}{Flat}
        & Diffusion Policy~\cite{chi2025diffusion} & 16 & \textcolor{lightgray}{/} & \textcolor{lightgray}{0} & \textcolor{lightgray}{/} & \textcolor{lightgray}{0} & \textcolor{lightgray}{/} & \textcolor{lightgray}{0} & \textcolor{lightgray}{/} & \textcolor{lightgray}{0} & \textcolor{lightgray}{/} & \textcolor{lightgray}{0} & \textcolor{lightgray}{/} & \textcolor{lightgray}{0} & \textcolor{lightgray}{/} & \textcolor{lightgray}{0} & \textcolor{lightgray}{/} & \textcolor{lightgray}{0} & \textcolor{lightgray}{/} \\
        & OpenVLA-OFT~\cite{kim24openvla}      & \textcolor{lightgray}{0} & \textcolor{lightgray}{/} & \textcolor{lightgray}{0} & \textcolor{lightgray}{/} & \textcolor{lightgray}{0} & \textcolor{lightgray}{/} & \textcolor{lightgray}{0} & \textcolor{lightgray}{/} & \textcolor{lightgray}{0} & \textcolor{lightgray}{/} & \textcolor{lightgray}{0} & \textcolor{lightgray}{/} & \textcolor{lightgray}{0} & \textcolor{lightgray}{/} & \textcolor{lightgray}{0} & \textcolor{lightgray}{/} & \textcolor{lightgray}{0} & \textcolor{lightgray}{/} \\
        & \(\pi_{0}\)~\cite{black2024pi_0}      & 12 & \textcolor{lightgray}{/} & \textcolor{lightgray}{0} & \textcolor{lightgray}{/} & \textcolor{lightgray}{0} & \textcolor{lightgray}{/} & \textcolor{lightgray}{0} & \textcolor{lightgray}{/} & \textcolor{lightgray}{0} & \textcolor{lightgray}{/} & \textcolor{lightgray}{0} & \textcolor{lightgray}{/} & \textcolor{lightgray}{0} & \textcolor{lightgray}{/} & \textcolor{lightgray}{0} & \textcolor{lightgray}{/} & \textcolor{lightgray}{0} & \textcolor{lightgray}{/} \\
        & \(\pi_{0.5}\)~\cite{intelligence2025pi_}    & 18 & \textcolor{lightgray}{/} & \textcolor{lightgray}{0} & \textcolor{lightgray}{/} & \textcolor{lightgray}{0} & \textcolor{lightgray}{/} & \textcolor{lightgray}{0} & \textcolor{lightgray}{/} & \textcolor{lightgray}{0} & \textcolor{lightgray}{/} & \textcolor{lightgray}{0} & \textcolor{lightgray}{/} & \textcolor{lightgray}{0} & \textcolor{lightgray}{/} & \textcolor{lightgray}{0} & \textcolor{lightgray}{/} & \textcolor{lightgray}{0} & \textcolor{lightgray}{/} \\
        \midrule
        \multirow{5}{*}{Modular}
        & RoboPoint~\cite{yuan2024robopoint} & \textcolor{lightgray}{0} & \textcolor{lightgray}{0} & \textcolor{lightgray}{0} & \textcolor{lightgray}{0} & \textcolor{lightgray}{0} & \textcolor{lightgray}{0} & \textcolor{lightgray}{0} & \textcolor{lightgray}{0} & \textcolor{lightgray}{0} & \textcolor{lightgray}{0} & \textcolor{lightgray}{0} & \textcolor{lightgray}{0} & \textcolor{lightgray}{0} & \textcolor{lightgray}{0} & \textcolor{lightgray}{0} & \textcolor{lightgray}{0} & \textcolor{lightgray}{0} & \textcolor{lightgray}{0} \\
        & HAMSTER~\cite{li2025hamster}   & \textcolor{lightgray}{0} & \textcolor{lightgray}{0} & \textcolor{lightgray}{0} & \textcolor{lightgray}{0} & \textcolor{lightgray}{0} & \textcolor{lightgray}{0} & \textcolor{lightgray}{0} & \textcolor{lightgray}{0} & \textcolor{lightgray}{0} & \textcolor{lightgray}{0} & \textcolor{lightgray}{0} & \textcolor{lightgray}{0} & \textcolor{lightgray}{0} & \textcolor{lightgray}{0} & \textcolor{lightgray}{0} & \textcolor{lightgray}{0} & \textcolor{lightgray}{0} & \textcolor{lightgray}{0} \\
        & AnyPlace~\cite{zhao2025anyplace}  & 12 & 12 & 42 & 42 & \textcolor{lightgray}{0} & \textcolor{lightgray}{0} & 38 & 38 & \textcolor{lightgray}{0} & \textcolor{lightgray}{0} & \textcolor{lightgray}{0} & \textcolor{lightgray}{0} & {16} & {16} & \textcolor{lightgray}{0} & \textcolor{lightgray}{0} & \textcolor{lightgray}{0} & \textcolor{lightgray}{0} \\

        & \cellcolor{yellow!30}\textbf{\name (Ours)} 
        & \cellcolor{yellow!30}\textbf{92}  & \cellcolor{yellow!30}\textbf{96}
        & \cellcolor{yellow!30}\textbf{80}  & \cellcolor{yellow!30}\textbf{82}
        & \cellcolor{yellow!30}\textbf{76}  & \cellcolor{yellow!30}\textbf{84}
        & \cellcolor{yellow!30}\textbf{96}  & \cellcolor{yellow!30}\textbf{98}
        & \cellcolor{yellow!30}\textbf{88}  & \cellcolor{yellow!30}\textbf{90}
        & \cellcolor{yellow!30}\textbf{100} & \cellcolor{yellow!30}\textbf{100}
        & \cellcolor{yellow!30}\textbf{88}  & \cellcolor{yellow!30}\textbf{94}
        & \cellcolor{yellow!30}\textbf{96}  & \cellcolor{yellow!30}\textbf{96}
        & \cellcolor{yellow!30}\textbf{90}  & \cellcolor{yellow!30}\textbf{94} \\
        & \cellcolor{blue!8}\textbf{\name (Oracle)} 
        & \cellcolor{blue!8}\textbf{98}  & \cellcolor{blue!8}\textbf{100}
        & \cellcolor{blue!8}\textbf{96}  & \cellcolor{blue!8}\textbf{100}
        & \cellcolor{blue!8}\textbf{82}  & \cellcolor{blue!8}\textbf{100}
        & \cellcolor{blue!8}\textbf{100}  & \cellcolor{blue!8}\textbf{100}
        & \cellcolor{blue!8}\textbf{98}  & \cellcolor{blue!8}\textbf{100}
        & \cellcolor{blue!8}\textbf{100} & \cellcolor{blue!8}\textbf{100}
        & \cellcolor{blue!8}\textbf{100}  & \cellcolor{blue!8}\textbf{100}
        & \cellcolor{blue!8}\textbf{100}  & \cellcolor{blue!8}\textbf{100}
        & \cellcolor{blue!8}\textbf{98}  & \cellcolor{blue!8}\textbf{100} \\
        \bottomrule[1.2pt]
    \end{tabular}
    }
    \label{tab:main_exp}
    
\end{table*}

\subsection{Main Results}

We compare \name and representative baselines: flat VLA, modular two-stage, and oracle baselines under the same zero-shot evaluation protocol. Following prior work~\cite{liu2023libero}, each method is evaluated with \(50\) trials per task. Results are reported in Tab.~\ref{tab:main_exp}, Fig.~\ref{fig:exp} and Fig.~\ref{fig:real_goal}.

\noindent\textbf{\ding{182} \name excels at slot-level placement.}
\name is the only method that achieves consistently high success across all nine SlotBench categories, from ordinal selection to comparison and logical negation.
In contrast, flat policies only show sparse success on the ordinal category, while modular baselines improve a few categories but remain unreliable overall.
\noindent\textbf{\ding{183} Spatial visual goals are crucial.}
VLM-based coordinate grounding is limited by localization precision, compositional reasoning, and generalization to unseen layouts.
Our image-generation-based goal construction instead produces spatially explicit visual targets, leading to substantially stronger IA and SR. \noindent\textbf{\ding{184} Oracle guidance reveals the bottleneck.}
The expert-human oracle achieves near-ceiling performance, showing that the low-level controller can execute accurately with correct targets.
Thus, the remaining gap mainly stems from high-level grounding errors, highlighting the importance of robust goal construction.

\setlength{\arrayrulewidth}{0.4pt} 
\renewcommand\arraystretch{1.4}
\begin{table*}[!t]
    \scriptsize
    \caption{\textbf{Ablation on high-level grounding module.} Report Coarse Accuracy (CA, \%), Instruction Accuracy (IA, \%) and Latency (s). $\dagger$ means inference performed on a H200 GPU. The yellow rows denote the module used by \name. \(\ddagger\) means fine-tuned and inference performed on a H200 GPU.} 
    \centering
    \scalebox{0.78}{
    \begin{tabular}{c c |
        p{0.45cm}<{\centering} p{0.45cm}<{\centering}|
        p{0.45cm}<{\centering} p{0.45cm}<{\centering}|
        p{0.45cm}<{\centering} p{0.45cm}<{\centering}|
        p{0.45cm}<{\centering} p{0.45cm}<{\centering}|
        p{0.45cm}<{\centering} p{0.45cm}<{\centering}|
        p{0.45cm}<{\centering} p{0.45cm}<{\centering}|
        p{0.45cm}<{\centering} p{0.45cm}<{\centering}|
        p{0.45cm}<{\centering} p{0.45cm}<{\centering}|
        p{0.45cm}<{\centering} p{0.45cm}<{\centering}|
        c}
        \toprule[1.2pt]
        \multirow{2}{*}{\textbf{Group}} & \multirow{2}{*}{\textbf{Method}} 
        & \multicolumn{2}{c|}{\textbf{Ord.}} & \multicolumn{2}{c|}{\textbf{Size}} 
        & \multicolumn{2}{c|}{\textbf{Hgt.}} & \multicolumn{2}{c|}{\textbf{Dist.}} 
        & \multicolumn{2}{c|}{\textbf{Comp.}} & \multicolumn{2}{c|}{\textbf{Neg.}} 
        & \multicolumn{2}{c|}{\textbf{Vague}} & \multicolumn{2}{c|}{\textbf{Aff.}} 
        & \multicolumn{2}{c|}{\textbf{World}} 
        & \multirow{2}{*}{\textbf{Latency}} \\
        \cmidrule{3-20}
        & & CA & IA & CA & IA & CA & IA & CA & IA & CA & IA & CA & IA & CA & IA & CA & IA & CA & IA & \\
        \cmidrule{1-21}
        \multirow{5}{*}{VLM}
        & Molmo-7B$^\dagger$    & 44 & 12  & 48 & 42& 6  & \textcolor{lightgray}{0}  & 38 & \textcolor{lightgray}{0} & 4 & \textcolor{lightgray}{0}  & \textcolor{lightgray}{0} & \textcolor{lightgray}{0} & \textcolor{lightgray}{0} & \textcolor{lightgray}{0} & \textcolor{lightgray}{0} & \textcolor{lightgray}{0} & \textcolor{lightgray}{0} & \textcolor{lightgray}{0} & 2 \\
        & GPT-5.3-instant & 8 & \textcolor{lightgray}{0} & 24 & 6 & \textcolor{lightgray}{0} & \textcolor{lightgray}{0} & 6 & \textcolor{lightgray}{0}  & 8 & 2  & 2 & \textcolor{lightgray}{0} & 4 & 2 & 12 & 2 & 16 & 4 & 25 \\
        & GPT-5.2 Think       & 52 & 34  & 80 & 72 & 52 & 28 & 82 & 40 & 62 & 16 & 78 & 62 & 66 & 40 & 48 & 42 & 72 & 46 & 133 \\
        & Gemini-3.0 Pro  & 80 & 74 & 78 & 52 & 48 & 38 & 56 & 50 & 38 & 24 & 100 & 78 & 82 & 60 & 88 & 76 & 78 & 32 & 89 \\
        & Gemini-3.1 Pro      & 98 & 92 & 90 & 84 & 86 & 60 & 100 & 44 & 92 & 48 & 100 & 72 & 100 & 76 & 100 & 92 & 100 & 62 & 121\\
        
        \midrule
        \multirow{5}{0.7in}{\centering Image \\ Generator}
        & Grok-Image & 58 & 58 & 42 & 42 & 54 & 54 & 94 & 94 & 24 & 24 & 32 & 32 & 88 & 88 & 38 & 38 & 88 & 88 & 9\\
        & GPT-Image-1.5   & \textcolor{lightgray}{0} & \textcolor{lightgray}{0} & \textcolor{lightgray}{0} & \textcolor{lightgray}{0} & \textcolor{lightgray}{0} & \textcolor{lightgray}{0} & \textcolor{lightgray}{0} & \textcolor{lightgray}{0} & \textcolor{lightgray}{0} & \textcolor{lightgray}{0} & \textcolor{lightgray}{0} & \textcolor{lightgray}{0} & \textcolor{lightgray}{0} & \textcolor{lightgray}{0} & \textcolor{lightgray}{0} & \textcolor{lightgray}{0} & \textcolor{lightgray}{0} & \textcolor{lightgray}{0} & 22 \\
        & Flux2-Dev\(\ddagger\)  & 12 & 12 & 20 & 20 & 30 & 30 & 30 & 30 & 10 & 10 & 20 & 20 & 30 & 30 & 20 & 20 & 40 & 40 & 4 \\
        & Stable Diffusion-XL\(\ddagger\)  & 10 & 10 & 10 & 10 & \textcolor{lightgray}{0} & \textcolor{lightgray}{0} & 20 & 20 & 20 & 20 & \textcolor{lightgray}{0} & \textcolor{lightgray}{0} & \textcolor{lightgray}{0} & \textcolor{lightgray}{0} & 10 & 10 & 20 & 20 & 1 \\
        & Nano-Banana Pro 
        & 72  & 72
        & 84  & 84
        & 72  & 72
        & 94  & 94
        & 92  & 92
        & 100 & 100
        & 88  & 88
        & 96  & 96
        & 92  & 92 & 16 \\
        & \cellcolor{yellow!30}\textbf{Nano-Banana 2} & \cellcolor{yellow!30}\textbf{96} & \cellcolor{yellow!30}\textbf{96} & \cellcolor{yellow!30}\textbf{82} & \cellcolor{yellow!30}\textbf{82} & \cellcolor{yellow!30}\textbf{84} & \cellcolor{yellow!30}\textbf{84} & \cellcolor{yellow!30}\textbf{98} & \cellcolor{yellow!30}\textbf{98} & \cellcolor{yellow!30}\textbf{90} & \cellcolor{yellow!30}\textbf{90} & \cellcolor{yellow!30}\textbf{100} & \cellcolor{yellow!30}\textbf{100} & \cellcolor{yellow!30}\textbf{94} & \cellcolor{yellow!30}\textbf{94} & \cellcolor{yellow!30}\textbf{96} & \cellcolor{yellow!30}\textbf{96} & \cellcolor{yellow!30}\textbf{94} & \cellcolor{yellow!30}\textbf{94} & \cellcolor{yellow!30}\textbf{11} \\
        \bottomrule[1.2pt]
    \end{tabular}
    }
    
    \label{tab:ablation1}
\end{table*}

\subsection{Ablation Study}

\stitle{Alternative high-level grounding module.}
We compare recent VLMs and image generators as high-level grounding modules. VLMs output a placement coordinate as the goal center while image generators produce an edited guidance image with a spatial marker to define the goal center. Tab.~\ref{tab:ablation1} shows that reasoning VLMs localize slots coarsely and yield moderate CA but substantially lower IA. This result indicates insufficient spatial fidelity for sub-centimeter placement, and latency is also high. In contrast, image generators achieve consistently strong CA and IA at much lower latency, which is critical for robotic manipulation. However, some generators are impractical. Models such as GPT-Image-1.5 enforce resizing during generation, which breaks pixel-level alignment.

\stitle{Alternative low-level policy.} We compare low-level policies for slot-level placement in Tab.~\ref{tab:ablation2}. We use oracle guidance to isolate low-level execution from grounding errors. All policies are trained on the same synthetic dataset $\mathcal{D}{\mathrm{syn}}$ under comparable protocols. Diffusion Policy~\cite{chi2025diffusion} succeeds only when the viewpoint distribution matches training and fails elsewhere, indicating that generic behavior cloning lacks the precision required for slot-level insertion. Parameter-efficient tuning methods, including OpenVLA-OFT and \(\pi_{0.5}\) (PEFT), also perform poorly, suggesting limited adaptation capacity is insufficient. Fully fine-tuned $\pi_{0.5}$ achieves high success rates across categories, demonstrating that reliable slot-level manipulation requires substantial low-level policy adaptation.

\setlength{\arrayrulewidth}{0.4pt} 
\renewcommand\arraystretch{1.4}
\begin{table*}[!t]
    \scriptsize
    \caption{\textbf{Effect of the low-level policy.} Success Rate (SR, \%) on SlotBench. The performance varies significantly with different policy choices, and the full fine-tuned $\pi_{0.5}$ policy achieves the best results. yellow rows denote our proposed \name.} 
    \centering
    \resizebox{0.8\textwidth}{!}{
    \begin{tabular}{c |
        p{0.9cm}<{\centering}|
        p{0.9cm}<{\centering}|
        p{0.9cm}<{\centering}|
        p{0.9cm}<{\centering}|
        p{0.9cm}<{\centering}|
        p{0.9cm}<{\centering}|
        p{0.9cm}<{\centering}|
        p{0.9cm}<{\centering}|
        p{0.9cm}<{\centering}}
        \toprule[1.2pt]
        \multirow{2}{*}{\textbf{Method}} 
        & \textbf{Ord.} & \textbf{Size} & \textbf{Hgt.} & \textbf{Dist.} & \textbf{Comp.} & \textbf{Neg.} & \textbf{Vague} & \textbf{Aff.} & \textbf{World} \\
        \cmidrule{2-10}
        & SR & SR & SR & SR & SR & SR & SR & SR & SR \\
        \cmidrule{1-10}

        \name-Diffusion Policy         & 92 & \textcolor{lightgray}{0} & \textcolor{lightgray}{0} & \textcolor{lightgray}{0} & 94 & 90 & 98 & \textcolor{lightgray}{0} & \textcolor{lightgray}{0} \\
        \name-OpenVLA-OFT      &\textcolor{lightgray}{0} & \textcolor{lightgray}{0} & \textcolor{lightgray}{0} & \textcolor{lightgray}{0} & \textcolor{lightgray}{0} & \textcolor{lightgray}{0} & \textcolor{lightgray}{0} & \textcolor{lightgray}{0} &
        \textcolor{lightgray}{0} \\
         \name-\(\pi_{0.5}\) (PEFT) & 10  & 12 & 4 & 8 & 12 & 8 & 2 & 0 & 10 \\
        \cellcolor{yellow!30}\textbf{\name-\(\pi_{0.5}\) (FT)(Baseline)} & \cellcolor{yellow!30}\textbf{98} & \cellcolor{yellow!30}\textbf{96} & \cellcolor{yellow!30}\textbf{82} & \cellcolor{yellow!30}\textbf{100} & \cellcolor{yellow!30}\textbf{100} & \cellcolor{yellow!30}\textbf{100} & \cellcolor{yellow!30}\textbf{100} & \cellcolor{yellow!30}\textbf{100} & \cellcolor{yellow!30}\textbf{98} \\

        \bottomrule[1.2pt]
    \end{tabular}
    }
    
    \label{tab:ablation2}
\end{table*}

\setlength{\arrayrulewidth}{0.4pt} 
\renewcommand\arraystretch{1.4}
\begin{table*}[!t]
    \scriptsize
    \caption{\textbf{Effect of the synthetic dataset.} Success Rate (SR, \%) on SlotBench. Training with $\mathcal{D}_{\mathrm{syn}}$ yields significantly higher SR than $\mathcal{D}_{\mathrm{std}}$ across all categories. yellow rows denote \name.} 
    \centering
    \resizebox{0.8\textwidth}{!}{
    \begin{tabular}{c |
        p{0.9cm}<{\centering}|
        p{0.9cm}<{\centering}|
        p{0.9cm}<{\centering}|
        p{0.9cm}<{\centering}|
        p{0.9cm}<{\centering}|
        p{0.9cm}<{\centering}|
        p{0.9cm}<{\centering}|
        p{0.9cm}<{\centering}|
        p{0.9cm}<{\centering}}
        \toprule[1.2pt]
        \multirow{2}{*}{\textbf{Method}} 
        & \textbf{Ord.} & \textbf{Size} & \textbf{Hgt.} & \textbf{Dist.} & \textbf{Comp.} & \textbf{Neg.} & \textbf{Vague} & \textbf{Aff.} & \textbf{World} \\
        \cmidrule{2-10}
        & SR & SR & SR & SR & SR & SR & SR & SR & SR \\
        \cmidrule{1-10}
         \name (w/ \(\mathcal{D}_{\mathrm{std}}\))       &2 & \textcolor{lightgray}{0} & \textcolor{lightgray}{0} & \textcolor{lightgray}{0} & \textcolor{lightgray}{0} & \textcolor{lightgray}{0} & \textcolor{lightgray}{0} & \textcolor{lightgray}{0} &
        \textcolor{lightgray}{0} \\
        \cellcolor{yellow!30}\textbf{\name (w/ \(\mathcal{D}_{\mathrm{syn}}\))(Baseline)} & \cellcolor{yellow!30}\textbf{98} & \cellcolor{yellow!30}\textbf{96} & \cellcolor{yellow!30}\textbf{82} & \cellcolor{yellow!30}\textbf{100} & \cellcolor{yellow!30}\textbf{100} & \cellcolor{yellow!30}\textbf{100} & \cellcolor{yellow!30}\textbf{100} & \cellcolor{yellow!30}\textbf{100} & \cellcolor{yellow!30}\textbf{98} \\

        \bottomrule[1.2pt]
    \end{tabular}
    }
    \label{tab:ablation3}
    
\end{table*}

\stitle{Training without synthetic dataset.}
We replace the synthetic dataset \(\mathcal{D}_{\mathrm{syn}}\), which includes a blue guidance sphere, with the standard dataset \(\mathcal{D}_{\mathrm{std}}\), which provides only language instructions, to quantify the benefit of synthetic visual-goal augmentation for low-level training. As shown in Tab.~\ref{tab:ablation3}, policies trained on \(\mathcal{D}_{\mathrm{std}}\) achieve near-zero success across tasks, indicating that synthetic visual-goal supervision is critical for learning precise slot-level execution.

\setlength{\arrayrulewidth}{0.4pt} 
\renewcommand\arraystretch{1.4}
\begin{table*}[!t]
    \scriptsize
    \caption{\textbf{Effect of in-domain training on SlotBench.} Success Rate (SR, \%). 
    Directly fine-tuning $\pi_{0.5}$ in-domain yields limited gains, while \name\ achieves substantially higher performance across all categories. yellow rows denote our proposed \name.}

    \centering
    \resizebox{0.8\textwidth}{!}{
    \begin{tabular}{c |
        p{0.9cm}<{\centering}|
        p{0.9cm}<{\centering}|
        p{0.9cm}<{\centering}|
        p{0.9cm}<{\centering}|
        p{0.9cm}<{\centering}|
        p{0.9cm}<{\centering}|
        p{0.9cm}<{\centering}|
        p{0.9cm}<{\centering}|
        p{0.9cm}<{\centering}}
        \toprule[1.2pt]
        \multirow{2}{*}{\textbf{Method}} 
        & \textbf{Ord.} & \textbf{Size} & \textbf{Hgt.} & \textbf{Dist.} & \textbf{Comp.} & \textbf{Neg.} & \textbf{Vague} & \textbf{Aff.} & \textbf{World} \\
        \cmidrule{2-10}
        & SR & SR & SR & SR & SR & SR & SR & SR & SR \\
        \cmidrule{1-10}
         \(\pi_{0.5}\) (in-domain)       &42 & 38 & 26 & 26 & 18 & 30 & 25 & 32 &
        22 \\
        \cellcolor{yellow!30}\textbf{\name (Baseline)} & \cellcolor{yellow!30}\textbf{98} & \cellcolor{yellow!30}\textbf{96} & \cellcolor{yellow!30}\textbf{82} & \cellcolor{yellow!30}\textbf{100} & \cellcolor{yellow!30}\textbf{100} & \cellcolor{yellow!30}\textbf{100} & \cellcolor{yellow!30}\textbf{100} & \cellcolor{yellow!30}\textbf{100} & \cellcolor{yellow!30}\textbf{98} \\

        \bottomrule[1.2pt]
    \end{tabular}
    }
    
    \label{tab:ablation4}
\end{table*}

\stitle{In-domain flat policy training.}
We fine-tune a flat policy on in-domain data from nine SlotBench tasks. For each task, we include at least two tray-shaped variants (of five total) and at least two instruction variants (of five total). As shown in Tab.~\ref{tab:ablation4}, the flat policy succeeds mainly on near-duplicates of the training distribution, but mild geometry or language changes cause sharp drops, suggesting overfitting. In contrast, \name remains robust under unseen layouts.

\subsection{Real-World Experiment}

We validate \name\ on a real UR10e platform with two RealSense D455 RGB-D cameras (Fig.~\ref{fig:real_goal}), using a red block and a \(2\times2\) tray. 
We collect 50 ordinal-reasoning demonstrations covering only two of the four slots, and evaluate on 20 trials with instructions covering all slots. 
\name\ achieves an 82\% success rate, showing that the visual goal can be generated, reconstructed, and executed in real-world settings while generalizing to unseen slot targets. 
More results are in Appendix.

\section{Conclusion}

We showed that slot-level placement remains challenging for current flat and VLM-based modular policies because flat policies entangle grounding with action, while modular pipelines suffer from imprecise intermediate grounding. We introduced AnySlot, a goal-conditioned framework that converts language into an explicit visual goal (via image generation) and executes it with a low-level policy, decoupling slot selection from precise placement. We also introduced SlotBench, a nine-task benchmark for evaluating structured slot-level reasoning under zero-shot transfer to unseen layouts and instructions. Across diverse settings, AnySlot consistently improves semantic correctness and geometric precision, demonstrating strong zero-shot generalization for slot-level manipulation.

\noindent\textbf{Limitation.} AnySlot has several limitations.
First, reliable goal generation currently depends on a strong external image-generation model. Although open-source alternatives were evaluated, they showed weaker spatial reasoning ability. Second, the zero-shot generalization mainly targets language grounding and unseen slot layouts, rather than general tasks and objects.
Finally, depth-based goal reconstruction may fail under extreme viewpoints, partial marker visibility or severe occlusion. 


\clearpage
\acknowledgments{If a paper is accepted, the final camera-ready version will (and probably should) include acknowledgments. All acknowledgments go at the end of the paper, including thanks to reviewers who gave useful comments, to colleagues who contributed to the ideas, and to funding agencies and corporate sponsors that provided financial support.}


\bibliography{example}  

@String(CVPR  = {IEEE Conf. Comput. Vis. Pattern Recog.})

@String(CVPR  = {CVPR})

@inproceedings{zitkovich2023rt,
  title={Rt-2: Vision-language-action models transfer web knowledge to robotic control},
  author={Zitkovich, Brianna and Yu, Tianhe and Xu, Sichun and Xu, Peng and Xiao, Ted and Xia, Fei and Wu, Jialin and Wohlhart, Paul and Welker, Stefan and Wahid, Ayzaan and others},
  booktitle={Conference on Robot Learning},
  pages={2165--2183},
  year={2023},
  organization={PMLR}
}

@article{kim24openvla,
    title={OpenVLA: An Open-Source Vision-Language-Action Model},
    author={{Moo Jin} Kim and Karl Pertsch and Siddharth Karamcheti and Ted Xiao and Ashwin Balakrishna and Suraj Nair and Rafael Rafailov and Ethan Foster and Grace Lam and Pannag Sanketi and Quan Vuong and Thomas Kollar and Benjamin Burchfiel and Russ Tedrake and Dorsa Sadigh and Sergey Levine and Percy Liang and Chelsea Finn},
    journal = {arXiv preprint arXiv:2406.09246},
    year={2024},
}

@article{black2024pi_0,
  title={$\pi_0 $: A Vision-Language-Action Flow Model for General Robot Control},
  author={Black, Kevin and Brown, Noah and Driess, Danny and Esmail, Adnan and Equi, Michael and Finn, Chelsea and Fusai, Niccolo and Groom, Lachy and Hausman, Karol and Ichter, Brian and others},
  journal={arXiv preprint arXiv:2410.24164},
  year={2024}
}

@article{intelligence2025pi_,
  title={$\pi_0.5$: a Vision-Language-Action Model with Open-World Generalization},
  author={Intelligence, Physical and Black, Kevin and Brown, Noah and Darpinian, James and Dhabalia, Karan and Driess, Danny and Esmail, Adnan and Equi, Michael and Finn, Chelsea and Fusai, Niccolo and others},
  journal={arXiv preprint arXiv:2504.16054},
  year={2025}
}

@article{shi2025hi,
  title={Hi robot: Open-ended instruction following with hierarchical vision-language-action models},
  author={Shi, Lucy Xiaoyang and Ichter, Brian and Equi, Michael and Ke, Liyiming and Pertsch, Karl and Vuong, Quan and Tanner, James and Walling, Anna and Wang, Haohuan and Fusai, Niccolo and others},
  journal={arXiv preprint arXiv:2502.19417},
  year={2025}
}

@inproceedings{huang2025roboground,
  title={Roboground: Robotic manipulation with grounded vision-language priors},
  author={Huang, Haifeng and Chen, Xinyi and Chen, Yilun and Li, Hao and Han, Xiaoshen and Wang, Zehan and Wang, Tai and Pang, Jiangmiao and Zhao, Zhou},
  booktitle={Proceedings of the Computer Vision and Pattern Recognition Conference},
  pages={22540--22550},
  year={2025}
}

@article{yuan2025seeing,
  title={From seeing to doing: Bridging reasoning and decision for robotic manipulation},
  author={Yuan, Yifu and Cui, Haiqin and Chen, Yibin and Dong, Zibin and Ni, Fei and Kou, Longxin and Liu, Jinyi and Li, Pengyi and Zheng, Yan and Hao, Jianye},
  journal={arXiv preprint arXiv:2505.08548},
  year={2025}
}

@article{huang2025thinkact,
  title={Thinkact: Vision-language-action reasoning via reinforced visual latent planning},
  author={Huang, Chi-Pin and Wu, Yueh-Hua and Chen, Min-Hung and Wang, Yu-Chiang Frank and Yang, Fu-En},
  journal={arXiv preprint arXiv:2507.16815},
  year={2025}
}

@article{yuan2024robopoint,
  title={Robopoint: A vision-language model for spatial affordance prediction for robotics},
  author={Yuan, Wentao and Duan, Jiafei and Blukis, Valts and Pumacay, Wilbert and Krishna, Ranjay and Murali, Adithyavairavan and Mousavian, Arsalan and Fox, Dieter},
  journal={arXiv preprint arXiv:2406.10721},
  year={2024}
}

@article{zhao2025anyplace,
  title={AnyPlace: learning generalized object placement for robot manipulation},
  author={Zhao, Yuchi and Bogdanovic, Miroslav and Luo, Chengyuan and Tohme, Steven and Darvish, Kourosh and Aspuru-Guzik, Al{\'a}n and Shkurti, Florian and Garg, Animesh},
  journal={arXiv preprint arXiv:2502.04531},
  year={2025}
}

@article{chi2025diffusion,
  title={Diffusion policy: Visuomotor policy learning via action diffusion},
  author={Chi, Cheng and Xu, Zhenjia and Feng, Siyuan and Cousineau, Eric and Du, Yilun and Burchfiel, Benjamin and Tedrake, Russ and Song, Shuran},
  journal={The International Journal of Robotics Research},
  volume={44},
  number={10-11},
  pages={1684--1704},
  year={2025},
  publisher={Sage Publications Sage UK: London, England}
}

@article{mees2022calvin,
  title={Calvin: A benchmark for language-conditioned policy learning for long-horizon robot manipulation tasks},
  author={Mees, Oier and Hermann, Lukas and Rosete-Beas, Erick and Burgard, Wolfram},
  journal={IEEE Robotics and Automation Letters},
  volume={7},
  number={3},
  pages={7327--7334},
  year={2022},
  publisher={IEEE}
}

@article{li2024simpler,
  title={Evaluating real-world robot manipulation policies in simulation},
  author={Li, Xuanlin and Hsu, Kyle and Gu, Jiayuan and Pertsch, Karl and Mees, Oier and Walke, Homer Rich and Fu, Chuyuan and Lunawat, Ishikaa and Sieh, Isabel and Kirmani, Sean and others},
  journal={arXiv preprint arXiv:2405.05941},
  year={2024}
}

@article{liu2023libero,
  title={Libero: Benchmarking knowledge transfer for lifelong robot learning},
  author={Liu, Bo and Zhu, Yifeng and Gao, Chongkai and Feng, Yihao and Liu, Qiang and Zhu, Yuke and Stone, Peter},
  journal={Advances in Neural Information Processing Systems},
  volume={36},
  pages={44776--44791},
  year={2023}
}

@inproceedings{karamcheti2024prismatic,
  title={Prismatic vlms: Investigating the design space of visually-conditioned language models},
  author={Karamcheti, Siddharth and Nair, Suraj and Balakrishna, Ashwin and Liang, Percy and Kollar, Thomas and Sadigh, Dorsa},
  booktitle={Forty-first International Conference on Machine Learning},
  year={2024}
}

@article{beyer2024paligemma,
  title={Paligemma: A versatile 3b vlm for transfer},
  author={Beyer, Lucas and Steiner, Andreas and Pinto, Andr{\'e} Susano and Kolesnikov, Alexander and Wang, Xiao and Salz, Daniel and Neumann, Maxim and Alabdulmohsin, Ibrahim and Tschannen, Michael and Bugliarello, Emanuele and others},
  journal={arXiv preprint arXiv:2407.07726},
  year={2024}
}

@article{steiner2024paligemma,
  title={Paligemma 2: A family of versatile vlms for transfer},
  author={Steiner, Andreas and Pinto, Andr{\'e} Susano and Tschannen, Michael and Keysers, Daniel and Wang, Xiao and Bitton, Yonatan and Gritsenko, Alexey and Minderer, Matthias and Sherbondy, Anthony and Long, Shangbang and others},
  journal={arXiv preprint arXiv:2412.03555},
  year={2024}
}

@inproceedings{radford2021learning,
  title={Learning transferable visual models from natural language supervision},
  author={Radford, Alec and Kim, Jong Wook and Hallacy, Chris and Ramesh, Aditya and Goh, Gabriel and Agarwal, Sandhini and Sastry, Girish and Askell, Amanda and Mishkin, Pamela and Clark, Jack and others},
  booktitle={International conference on machine learning},
  pages={8748--8763},
  year={2021},
  organization={PmLR}
}

@inproceedings{deitke2025molmo,
  title={Molmo and pixmo: Open weights and open data for state-of-the-art vision-language models},
  author={Deitke, Matt and Clark, Christopher and Lee, Sangho and Tripathi, Rohun and Yang, Yue and Park, Jae Sung and Salehi, Mohammadreza and Muennighoff, Niklas and Lo, Kyle and Soldaini, Luca and others},
  booktitle={Proceedings of the Computer Vision and Pattern Recognition Conference},
  pages={91--104},
  year={2025}
}

@inproceedings{shukor2025scaling,
  title={Scaling laws for native multimodal models},
  author={Shukor, Mustafa and Fini, Enrico and da Costa, Victor Guilherme Turrisi and Cord, Matthieu and Susskind, Joshua and El-Nouby, Alaaeldin},
  booktitle={Proceedings of the IEEE/CVF International Conference on Computer Vision},
  pages={12--23},
  year={2025}
}

@inproceedings{yang2025thinking,
  title={Thinking in space: How multimodal large language models see, remember, and recall spaces},
  author={Yang, Jihan and Yang, Shusheng and Gupta, Anjali W and Han, Rilyn and Fei-Fei, Li and Xie, Saining},
  booktitle={Proceedings of the Computer Vision and Pattern Recognition Conference},
  pages={10632--10643},
  year={2025}
}

@article{li2025hamster,
  title={Hamster: Hierarchical action models for open-world robot manipulation},
  author={Li, Yi and Deng, Yuquan and Zhang, Jesse and Jang, Joel and Memmel, Marius and Yu, Raymond and Garrett, Caelan Reed and Ramos, Fabio and Fox, Dieter and Li, Anqi and others},
  journal={arXiv preprint arXiv:2502.05485},
  year={2025}
}

@inproceedings{KITE2023,
  author       = {Priya Sundaresan and
                  Suneel Belkhale and
                  Dorsa Sadigh and
                  Jeannette Bohg},
  editor       = {Jie Tan and
                  Marc Toussaint and
                  Kourosh Darvish},
  title        = {{KITE:} Keypoint-Conditioned Policies for Semantic Manipulation},
  booktitle    = {Conference on Robot Learning, CoRL 2023, 6-9 November 2023, Atlanta,
                  GA, {USA}},
  series       = {Proceedings of Machine Learning Research},
  volume       = {229},
  pages        = {1006--1021},
  publisher    = {{PMLR}},
  year         = {2023},
  bibsource    = {dblp computer science bibliography, https://dblp.org}
}

@InProceedings{Xiang_2020_SAPIEN,
author = {Xiang, Fanbo and Qin, Yuzhe and Mo, Kaichun and Xia, Yikuan and Zhu, Hao and Liu, Fangchen and Liu, Minghua and Jiang, Hanxiao and Yuan, Yifu and Wang, He and Yi, Li and Chang, Angel X. and Guibas, Leonidas J. and Su, Hao},
title = {{SAPIEN}: A SimulAted Part-based Interactive ENvironment},
booktitle = {The IEEE Conference on Computer Vision and Pattern Recognition (CVPR)},
month = {June},
year = {2020}}

@article{chen2025robotwin,
        title={RoboTwin 2.0: A Scalable Data Generator and Benchmark with Strong Domain Randomization for Robust Bimanual Robotic Manipulation},
        author={Chen, Tianxing and Chen, Zanxin and Chen, Baijun and Cai, Zijian and Liu, Yibin and Liang, Qiwei and Li, Zixuan and Lin, Xianliang and Ge, Yiheng and Gu, Zhenyu and others},
        journal={arXiv preprint arXiv:2506.18088},
        year={2025}
      }

@article{shan2025slot,
  title={Slot-Level Robotic Placement via Visual Imitation from Single Human Video},
  author={Shan, Dandan and Mo, Kaichun and Yang, Wei and Chao, Yu-Wei and Fouhey, David and Fox, Dieter and Mousavian, Arsalan},
  journal={arXiv preprint arXiv:2504.01959},
  year={2025}
}

@article{zhao2023learning,
  title={Learning fine-grained bimanual manipulation with low-cost hardware},
  author={Zhao, Tony Z and Kumar, Vikash and Levine, Sergey and Finn, Chelsea},
  journal={arXiv preprint arXiv:2304.13705},
  year={2023}
}

\end{document}